\documentclass{article}
\usepackage{spconf,amsmath,graphicx}

\usepackage{titletoc}
\usepackage{xcolor}
\definecolor{toccolor}{RGB}{0,0,255} 
\newcommand\DoToC{%
  \startcontents
  \titlecontents{section}[0pt]{\bfseries}{\thecontentslabel:\quad}{}{\hspace{1em}\titlerule*[1em]{.}\contentspage}
  \titlecontents{subsection}[16pt]{\bfseries}{\thecontentslabel:\quad}{}{\hspace{1em}\titlerule*[1em]{.}\contentspage}
  \printcontents{}{1}{\hrulefill\vskip1pt}
  \vskip0pt \noindent\hrulefill 
  \vspace{15pt}
  }

\usepackage{booktabs}
\usepackage{tabu}

\usepackage[pagebackref,breaklinks,colorlinks]{hyperref}

\usepackage[utf8]{inputenc} 
\usepackage[T1]{fontenc}    
\usepackage{url}            
\usepackage{booktabs}       
\usepackage{amsfonts}       
\usepackage{nicefrac}       
\usepackage{microtype}      
\usepackage{xcolor}         

\usepackage{bbding}

\usepackage{multirow}
\usepackage{subcaption}
\usepackage{bigstrut}
\usepackage{graphicx}
\usepackage{amsmath}
\usepackage{amssymb}
\usepackage{algorithm}
\usepackage{algorithmicx}
\usepackage{algpseudocode}
\usepackage{tabu}
\usepackage{siunitx}
\usepackage{adjustbox}
\usepackage{subcaption}
\usepackage{siunitx}
\usepackage{makecell}
\usepackage{xcolor}
\usepackage{colortbl}
\usepackage{color}
\usepackage{multirow}
\usepackage{blindtext}
\usepackage{comment}
\usepackage{bm}
\usepackage{bbm}
\usepackage{booktabs}
\usepackage{wrapfig}
\usepackage{array} 
\definecolor{Gray}{gray}{0.95}
\definecolor{Cyan}{rgb}{0.88,1,1}
\definecolor{Aliceblue}{rgb}{0.94, 0.97, 1.0}

\usepackage{color}

\title{Language-free Compositional Action Generation \\
via Decoupling Refinement}
\name{
Xiao Liu$^{1,5}$,  Guangyi Chen$^{2,3}$, Yansong Tang$^{1}$, Guangrun Wang$^{4,*}$, Xiao-Ping Zhang$^{1}$, Ser-Nam Lim$^{6}$
}
\address{
$^{1}$Shenzhen Key Laboratory of Ubiquitous Data Enabling,
\\Tsinghua Shenzhen International Graduate School, Tsinghua University, China \\
$^{2}$Carnegie Mellon University, Pittsburgh PA, USA \\
$^{3}$Mohamed bin Zayed University of Artificial Intelligence, Abu Dhabi, UAE \\
$^{4}$University of Oxford, UK 
$^{5}$Eindhoven University of Technology, NL 
$^{6}$University of Central Florida , USA \\
}
\begin{document}
\topmargin=0mm
\ninept
\maketitle
\renewcommand{\thefootnote}{\fnsymbol{footnote}}
\footnotetext[1]{Corresponding author.}
\renewcommand{\thefootnote}{\arabic{footnote}}

\begin{abstract}
Composing simple actions into complex actions is crucial yet challenging. Existing methods largely rely on language annotations to discern composable latent semantics, which is costly and labor-intensive.
In this study, we introduce a novel framework to generate compositional actions without language auxiliaries. Our approach consists of three components: Action Coupling, Conditional Action Generation, and Decoupling Refinement. Action Coupling 
integrates two sub-actions 
to generate pseudo-training examples.
Then, a conditional generative model, CVAE is employed to facilitate the diverse generation.
Decoupling Refinement leverages a self-supervised pre-trained model MAE to ensure semantic consistency between sub-actions and compositional actions.
Due to the lack of existing datasets containing both sub-actions and compositional actions, we create two new datasets, named HumanAct-C and UESTC-C.
Both qualitative and quantitative assessments are conducted to show our efficacy. \footnote[1]{Our code can be found at: \href{https://github.com/XLiu443/Language-free-Compositional-Action-Generation-via-Decoupling-Refinement}{https://github.com/XLiu443/Language-free-Compositional-Action-Generation-via-Decoupling-Refinement}}

\end{abstract}
\begin{keywords}
compositional generation, 3D action generation
\end{keywords}
\renewcommand{\thefootnote}{\fnsymbol{footnote}}


\section{Introduction}
\vspace{-2mm}

\label{sec:intro}
Humans possess the remarkable ability to generate new knowledge from prior experiences~\cite{lake2017building}. This capacity for compositional generation 
plays an indispensable role in artificial intelligence. This paper presents our efforts towards facilitating these compositional generation capabilities in 3D action generation. 
Although substantial strides have been made in synthesizing realistic 3D motion sequences~\cite{actor,guo2020action2motion,arad2021compositional}, generating complex 3D actions from basic elements presents its own unique challenges. This complexity arises from the difficulty in understanding the elemental latent concepts and the principles of compositions without any forms of supervision.

\begin{figure}[t]
\centering
\includegraphics[width=0.4\textwidth]{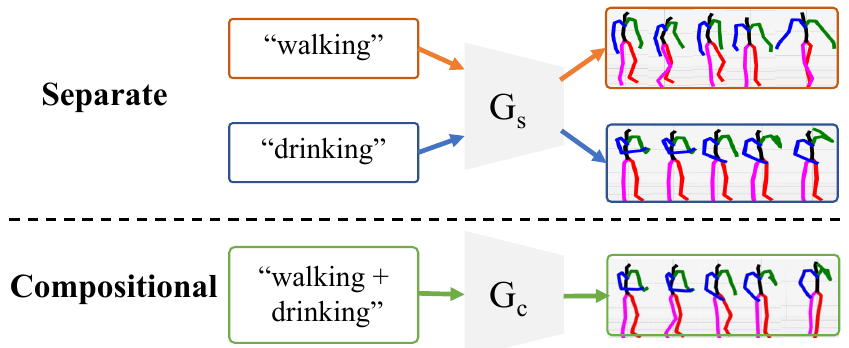}
\caption{ The comparison between separate action generation and compositional action generation. 
The compositional method aims to amalgamate two sub-actions, like ``walking'' and ``drinking'', into a simultaneous, unseen composite concept
- ``walking + drinking''. 
}
\label{fig:top1}
\end{figure}

Recent developments~\cite{motionclip,motiondiffuse,zhang2023remodiffuse} generate actions conditioned by text to reconcile the semantic space of language with the space of motion. MotionCLIP~\cite{motionclip} operates by positioning semantically analogous motions within the CLIP~\cite{clip} latent space into close proximity. MotionDiffuse~\cite{motiondiffuse} deploys CLIP for partial model initialization and progresses to learn a text-conditional diffusion model utilizing language annotations.
However, these methods have notable limitations. Firstly, they depend on an exhaustive collection of composite actions and corresponding high-quality text annotations, a process that is not only costly but also labor-intensive. Secondly, the composite actions gathered in this manner are predominantly temporally structured, like "walk and then run." As a result, they fall short in generating simultaneous compositional actions such as "walking while drinking."

In this paper, we consider a scenario that is challenging yet commonplace, where neither compositional action data nor annotated textual descriptions are provided and simultaneous compositions are involved. To surmount these obstacles, we propose a new framework aimed at learning the compositional action generation model using sub-actions solely. This framework is constituted of three components: \emph{Action Coupling}, \emph{Conditional Action Generation}, and \emph{Decoupling Refinement}.
\begin{itemize}
\setlength{\itemsep}{1pt}
\setlength{\parsep}{1pt}
\setlength{\parskip}{1pt}
  \item \emph{Action Coupling} is motivated by our observation that humans often concurrently perform two actions using separate body parts, such as ``drinking while walking'' in Figure~\ref{fig:top1}. 
  Thus, we propose 
  to couple two sub-actions that involve different active body parts serving as the pseudo-compositional actions. 
  \item \emph{Conditional Action Generation} is predicated on a conditional generative model, which uses a pair of sub-action labels as its conditioning variables. In particular, we utilize a Conditional Variational Autoencoder (CVAE)~\cite{cvae} model to develop a latent space for motion generation. 
  \item \emph{Decoupling Refinement} is introduced to improve the quality of generated actions, aiming to ensure semantic consistency between sub-actions and compositional actions. This is particularly crucial given that training data may become distorted during the coupling process. Specifically, we render and split the generated 3D skeletons into two sub-segments and use a pre-trained MAE model to recover the missing sections.
\end{itemize}
Owing to the absence of existing datasets that comprise both sub-actions and compositional actions, we devise two new datasets as benchmarks constructed on the HumanAct12~\cite{guo2020action2motion} and UESTC~\cite{ji2018large} datasets.
In these datasets, the training sets only consist of sub-actions, while the testing sets are assembled by manually applying an attention mask to each pair of sub-actions to compose them. 
Comprehensive quantitative and qualitative evaluations on two datasets substantiate the effectiveness of our proposed methodology in addressing the challenges of compositional action sequence generation.


\section{Related Work}
\label{sec: related}
\textbf{Human Motion Generation} 
Human motion generation has potential applications in fields requiring virtual or robotic characters. The primary methods employed in this domain can be categorized into generative adversarial network (GAN) based methods~\cite{gan}, flow-based methods~\cite{rezende2015variational}, and variational autoencoder (VAE) frameworks~\cite{kingma2013auto}. Notably, BAN~\cite{zhao2020bayesian} and Kinetic-GAN~\cite{degardin2022generative} apply GAN to model human motions while 
Action2motion~\cite{guo2020action2motion} and ACTOR~\cite{actor} propose different VAE-based architectures.
Recent methods~\cite{motionclip,motiondiffuse,zhang2023remodiffuse} have proposed using neural language as the conditional factor, which leverages the semantic  structure inherent in language to empower generation ability. 
Unlike these methods, our work targets a complex task—generating simultaneous compositional actions using minimal annotations, a paradigm that current methods don't address effectively.


\textbf{Compositional Image Generation.} In the field of image generation, there have been some developments to equip models with compositional capability. One prevalent approach involves learning disentangled factors of variation~\cite{nie2021controllable,wu2022object}. This method decomposes the observed object into distinct latent factors and then recombines these factors to generate new concepts. 
Another method focuses on learning a disentangled latent space in domain translation~\cite{benaim2019domain,press2020emerging}.
Other approaches employ large-scale language pre-trained models~\cite{clip,ramesh2021zero,dalle2} to achieve compositional image generation, but they often require a massive amount of data, running into the millions or even billions.



\section{Framework}

\begin{figure*}[t]
    \begin{center}
\centerline{\includegraphics[width=1.\linewidth]{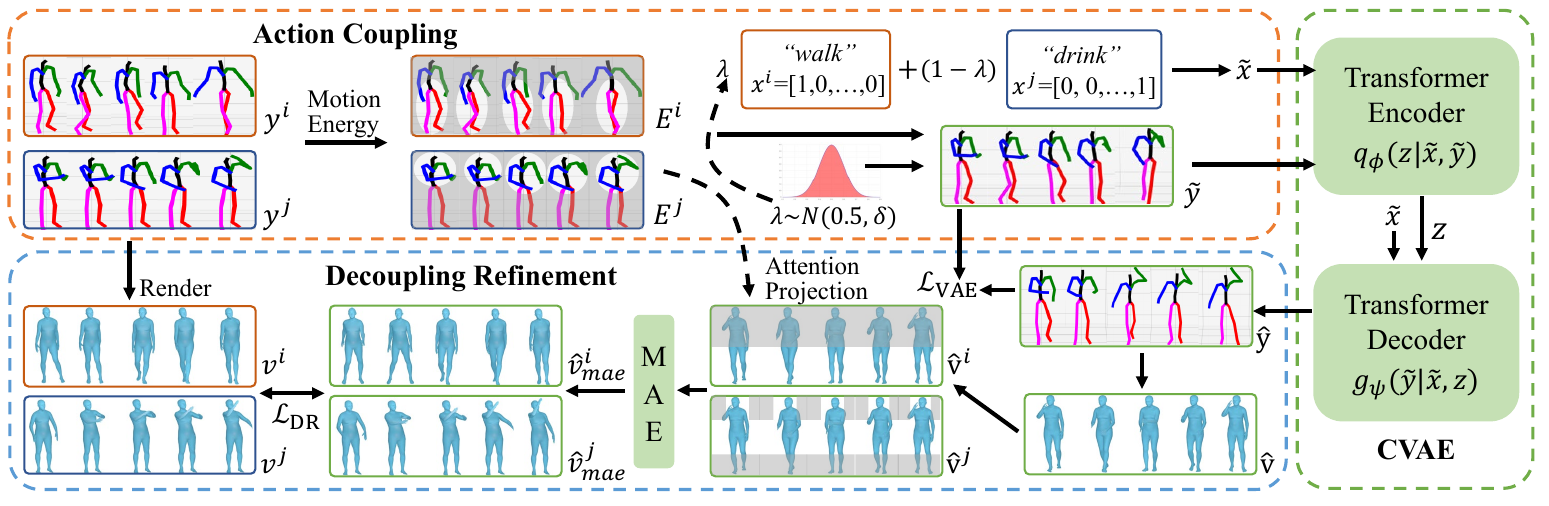}}
    \caption{ The pipeline of our framework. It contains three main components, including Action Coupling (in {\color{orange}orange}), Conditional Action Generation(in {\color{green}green}), and Decoupling Refinement (in {\color{blue}blue}). 
    These processes involve identifying active regions of sub-actions with motion energy, 
    and mixing sub-actions as pseudo-compositional actions for training a conditional generation model.    
    Generated compositional actions are then converted 
    and decoupled into masked images. A pre-trained MAE model is used to recover these images.
    }
\label{fig:framework}
    \end{center}
\end{figure*}

\subsection{Problem Definition}
We denote a pair of action labels and their corresponding motion sequences as $\mathbf{x},\mathbf{y} \in \mathcal{X}\times\mathcal{Y} $. Here, $\mathbf{x}$ signifies the sub-action label, 
while $\mathbf{y} =\{\mathbf{y}_t\in \mathbb{R}^{N\times3}| t=1,2,\cdots, T \} $ represents the sequence of action poses. Specifically, each frame $\mathbf{y}_t$ comprises a pose that includes N points $\{\textbf{y}_{n,t}| n =1,2,...,N\} $, each of which is a 3D coordinate, with $T$ denoting the sequence length.
Beyond this sub-action sequence, there lies an unseen compositional action space $\mathcal{Y}\times\mathcal{Y}\rightarrow\mathcal{\tilde{Y}}$, wherein each composite action is constructed from two sub-actions. The aim is to learn a generative model $g_{\psi}(\mathbf{\tilde{y}}|\mathbf{x}^i, \mathbf{x}^j,\mathbf{z})$ that produces a composite action $\mathbf{\tilde{y}} \in \mathcal{\tilde{Y}}$ given two different sub-action labels $\mathbf{x}^i, \mathbf{x}^j$ (here we use the superscript to represent different samples) and a noise vector $\mathbf{z}$.
\vspace{-1mm}

\subsection{Action Coupling}
\vspace{-1mm}
Action Coupling aims to synthesize compositional action sequences from observed sub-actions. 
Given two label variables $\mathbf{x}^i, \mathbf{x}^j $ and corresponding sub-action sequences $\mathbf{y}^i, \mathbf{y}^j $, we couple two sub-action sequences to generate pseudo targets in the following procedure: 
 \begin{equation}
  \begin{aligned}
\label{eq: mix_up} \tilde{\mathbf{x}} &= \lambda f_x^i(\mathbf{x}^i) + (1-\lambda) f_x^j(\mathbf{x}^j), \\
\tilde{\mathbf{y}} &= \lambda f_y^j(\mathbf{y}^j) + (1-\lambda) f_y^j(\mathbf{y}^j),
  \end{aligned}
\end{equation}
where the coefficient $\lambda \in [0,1]$ denotes the mixing rate between two source sub-actions, and functions $f_x, f_y$ represent the coupling strategy. In the following, we will discuss how to design the distribution of $\lambda$ and the implementation of functions $f_x, f_y$.

\textbf{Mixing rate distribution:}
In the context of Mix-up methods for data augmentation~\cite{zhang2018mixup}, the parameter $\lambda$ is sampled from $\lambda \sim Beta(\alpha,\alpha)$, where $\alpha \in (0,\inf)$ is a hyper-parameter that regulates the intensity of interpolation. This Beta distribution pushes the $\lambda$ close to $0$ or $1$ and away from $0.5$. It is reasonable, as overly mixed images (e.g., where $\lambda=0.5$) may dilute discriminative capacity, thereby contradicting the objective of data augmentation, which is to enhance classification.
Conversely, our method is centered around generating composite actions that preserve as much information from sub-actions as possible. In essence, we promote the deep integration of two actions, ideally where $\lambda =0.5$. As such, we employ $\lambda \sim \mathcal{N}(0.5,\delta)$, where $\delta$ denotes the variance of the Gaussian distribution.


\textbf{Preliminary insights:} Prior to exploring the mechanics of $f_x, f_y$, we aim to convey observations about human actions that serve as the motivation for our following designs. Firstly, it is important to recognize that \emph{a single body part is incapable of executing two distinct actions simultaneously}, 
which implies that simultaneous compositional actions generally involve two different body parts carrying out different sub-actions. This insight motivates our strategy of crafting two attention masks focusing on disparate body parts during action coupling.
Secondly, we find that \emph{body parts with the highest motion energy tend to be crucial in defining a particular action}. Essentially, significant motion is almost always associated with a body part that plays a key role in a certain sub-action.
This insight drives us to calculate motion energy to identify the active body parts.

\textbf{Energy-based attention masks:} Informed by the insights above, we formulate the subsequent energy model to compute the motion energy for each individual body part:
 \begin{equation}
  \begin{aligned}
\label{eq: motion energy} 
\mathbf{E}_{n} = \frac{1}{|\Omega|(T-1)}\sum_{\mathbf{y}_n\in \Omega }\sum_{t=2}^T ||\mathbf{y}_{n,t} - \mathbf{y}_{n,t-1}||^2,
  \end{aligned}
\end{equation}
where $\Omega$ denotes a body part, 
containing a set of joint points $\{\textbf{y}_{n,t}| n \in \Omega \}$, and $n,t$ denotes the $n$th joint point in the $t$th frame. Note that for the joint points in the same body part, the energy values are equal.
Then, we define the attention value as the energy value. 
Finally, we couple two sub-actions with the obtained attention mask $\mathbf{E}_{n}$ and implement functions $f_y$ in the second line in Eq~\eqref{eq: mix_up} as
 \begin{equation}
  \begin{aligned}
\label{eq: mix_up2} 
\tilde{\mathbf{y}}_n &= \frac{\lambda\textbf{E}^{i}_{n}\mathbf{y}^{i}_{n}}{\lambda\textbf{E}^{i}_{n} + (1-\lambda) \textbf{E}^{j}_{n}} + \frac{(1-\lambda) \textbf{E}^{j}_{n}\mathbf{y}^{j}_n}{\lambda\textbf{E}^{i}_{n} + (1-\lambda) \textbf{E}^{j}_{n}}
.
  \end{aligned}
\end{equation}
While $f_x$ is implemented by the identity function, i.e., $ \tilde{\mathbf{x}} = \lambda \mathbf{x}^i + (1-\lambda) \mathbf{x}^j$.

\subsection{Conditional Action Generation} 
Given the generated pseudo compositional training data $\tilde{\mathbf{y}}$ and $\tilde{\mathbf{x}}$, we formulate the generative process as $p(\tilde{\mathbf{y}}|\tilde{\mathbf{x}}) $:
 \begin{equation}
  \begin{aligned}
\label{eq: prediction_process} 
p(\tilde{\mathbf{y}}|\tilde{\mathbf{x}}) = \int g_{\psi}(\tilde{\mathbf{y}}|\tilde{\mathbf{x}},\mathbf{z})p_{\theta}(\mathbf{z}|\tilde{\mathbf{x}})d{\mathbf{z}}.
  \end{aligned}
\end{equation}
This generative process can be learned with CVAE
where a variational neural network $q_{\phi}(\mathbf{z}|\tilde{\mathbf{x}},\tilde{\mathbf{y}}) $
is introduced to optimize the objective above. The optimization follows the standard variational lower bound:
\begin{equation}
  \begin{aligned}
\label{eq: cvae_loss} \mathcal{L}_{CVAE}(\psi,\phi,\theta) =   \mathcal{L}_{KL}\big(q_{\phi}(\mathbf{z}|\tilde{\mathbf{x}},\tilde{\mathbf{y}})||p_{\theta}(\mathbf{z}|\tilde{\mathbf{x}})\big) \\
-\lambda \mathbb{E}_{q_{\phi}(\mathbf{z}|\tilde{\mathbf{x}},\tilde{\mathbf{y}})}[\log g_{\psi}(\tilde{\mathbf{y}}|\tilde{\mathbf{x}},\mathbf{z})].  
  \end{aligned}
\end{equation}
The first term reduces the KL divergence between $q_{\phi}(\mathbf{z}|\tilde{\mathbf{x}},\tilde{\mathbf{y}}) $ and  $p_{\theta}(\mathbf{z}|\tilde{\mathbf{x}})$, which encourages the consistency of latent variables generated from condition and data. The second term is a reconstruction objective that encourages the model to generate realistic action poses.

\subsection{ Decoupling Refinement}
Considering the potential compromise in the quality of generated sequences, 
we introduce \emph{Decoupling Refinement}, which
serves as a constraint to enhance the quality of the generated actions. 

\textbf{3D to 2D rendering:} In this study, we utilize the pre-trained MAE~\cite{mae} as our self-supervised model. 
However, there is a gap between 3D actions and 2D images (used for MAE). To bridge this gap, we utilize the Skinned Multi-Person Linear (SMPL) body model~\cite{loper2015smpl}.  Both the original sub-actions and the generated sequences are processed through the SMPL model, producing the dense point clouds representing 
body surfaces. These 3D outputs are then projected onto a 2D plane using a rendering function. This process can be formulated as  $f_{R} : \{\textbf{y}^i, \textbf{y}^j, \hat{\textbf{y}} \in \mathbb{R}^{3\times N}\} \rightarrow \{\textbf{v}^i, \textbf{v}^j, \hat{\textbf{v}} \in \mathbb{R}^{3\times H \times W}\}$, where $\textbf{v}$ denotes the rendering images with height $H$ and width $W$.
To enhance the readability of the resultant 2D images, we standardize the 3D model to ensure rendering is operated from a frontal perspective.

\textbf{Decoupling:} 
We introduce a reversal of the coupling process, referred as decoupling. 
which allows us to deconstruct the rendering images of compositional actions back to sub-action components and further facilitate refining generation. 
Intuitively, for a compositional action,
our goal is to preserve the active parts associated with each sub-action. 
Technically, we propose a 3D-2D attention projection to compute the decoupling attention map using the motion energy $\textbf{E}_n$. 
By utilizing the SMPL projection and the rendering function $f_R$,  we first transform each 3D joint point $\textbf{y}^i_n $ into 2D coordinates $\textbf{pix}^i_n $. 
The pixels in proximity to active joint points should be prioritized to focus. Hence, we establish a decay rate that is inversely proportional to the square of the pixel distance. We then compute the value for each pixel in the 2D attention map $\mathcal{A}$ by aggregating these decayed attention scores from all $N$ joint points:
 \begin{equation}
  \begin{aligned}
\label{eq: attention} 
\mathcal{A}^i(\textbf{pix}) = \sum_{n=1}^{N} \frac{\textbf{E}^i_n}{||\textbf{pix}-\textbf{pix}^i_n||^2},
  \end{aligned}
\end{equation}
where $\textbf{pix}$ denotes each pixel of the rendering image $\textbf{v}$. Lastly, we divide the image into segments of $16\times16$ regions, following the MAE model's configuration. We then compute the average attention value for each region and retain a subset of regions (e.g., 1/3) with high attention and masked the others. With this process, we obtain the decoupled images $\hat{\textbf{v}}^i=\hat{\textbf{v}} \circ \mathcal{A}^i $ 
and $\hat{\textbf{v}}^j=\hat{\textbf{v}} \circ \mathcal{A}^j$ from the compositional image.

\textbf{Constraints:} The masked images are fed to a pre-trained MAE model for the inpainting process. Despite the input image being part of a generated composite action, we encourage the model to reconstruct the original sub-actions, denoted as $\hat{\textbf{v}}_{mae}^i,\hat{\textbf{v}}_{mae}^j = f_{mae}(\hat{\textbf{v}}^i,\hat{\textbf{v}}^j )$, where $f_{mae}$ denotes the MAE model. It indicates that the attributes of the sub-actions are preserved during the compositional generation process and can be readily disentangled. For all data pairs $\textbf{y}^i,\textbf{y}^j$, the loss function for the decoupling refinement process can be expressed as follows:
 \begin{equation}
  \begin{aligned}
\label{eq: mae_loss} 
\mathcal{L}_{DR}= \sum_{i,j} (||\hat{\textbf{v}}_{mae}^i-\textbf{v}^i||^2 +||\hat{\textbf{v}}_{mae}^j-\textbf{v}^j||^2) .
  \end{aligned}
\end{equation}
Finally, the CVAE loss $\mathcal{L}_{CVAE}$ and the Decoupling Refinement loss $\mathcal{L}_{DR}$ are integrated to train our model.

\begin{table*}
\centering
\caption{Comparison with baseline methods on UESTC-C and HumanAct-C. Lower FID is better. Acc denotes the action recognition accuracy (higher is better). Div and Mul respectively denote the overall and per-class diversity. }
\setlength{\tabcolsep}{6pt}
\resizebox{.99\linewidth}{!}{
\begin{tabular}{lccrc|cccc}
    \toprule
        & \multicolumn{4}{c|}{UESTC-C} & \multicolumn{4}{c}{ HumanAct-C} \\
    Metrics 
    & FID$\downarrow$ & \multicolumn{1}{c}{Acc.$\uparrow$} & Div.$\rightarrow$ & Multimod.$\rightarrow$ & FID$\downarrow$ & Acc.$\uparrow$ & Div.$\rightarrow$ & Multimod.$\rightarrow$ \\
        \midrule
          Real 
          & $0.00^{\pm0.00}$ & $100.0^{\pm0.00}$ & $38.52^{\pm0.76}$ & $5.03^{\pm0.00}$ & $0.00^{\pm0.00}$ & $100.0^{\pm0.00}$ & $30.09^{\pm0.73}$ & $6.79^{\pm0.06}$\\
        \midrule
    Pre-train+Mask~\cite{actor} & $164.17 ^{\pm47.51}$ & $67.80 ^{\pm3.13}$ & $28.79^{\pm0.42}$ & $12.24^{\pm0.23}$  & $162.35^{\pm34.72}$ & $49.7^{\pm2.95}$ & $21.74^{\pm0.60}$ & $14.35^{\pm0.52}$   \\
   Latent Disentanglement~\cite{arad2021compositional} & $121.27 ^{\pm18.05}$ & $83.20^{\pm3.48}$ & $32.69 ^{\pm0.63}$ & $ 9.50^{\pm0.20}$    &
   \textbf{ $123.08^{\pm28.57}$} & \textbf{$52.2^{\pm4.84}$} & \textbf{$21.0^{\pm0.21}$} & \textbf{$14.79^{\pm0.53}$} \\

    MotionCLIP~\cite{motionclip}     & \textbf{$2502.81^{\pm0.00}$} &  \textbf{$4.00^{\pm0.00}$}& \textbf{$15.00^{\pm0.34}$} &  \textbf{$0.00^{\pm0.00}$}&\textbf{$6908.91^{\pm0.00}$}  & \textbf{$11.1^{\pm0.00}$} & \textbf{$14.81^{\pm0.63}$} &\textbf{$0.00^{\pm0.00}$} \\
\hline
\rowcolor{Aliceblue}
Ours(w/ DR) & \textbf{$90.13^{\pm12.58}$ } & \textbf{$87.20^{\pm1.69}$} & \textbf{$33.06^{\pm0.81}$} & \textbf{$10.12^{\pm0.06}$} 
& \textbf{$83.43^{\pm13.01}$ } & \textbf{$64.1^{\pm5.41}$} & \textbf{$21.36^{\pm0.39}$} & \textbf{$14.01^{\pm0.38}$}\\
        \bottomrule
\end{tabular}
}
\label{tab:comparison}
\end{table*}

\begin{table}\tiny
\centering
\caption{Ablation studies on UESTC-C and HumanAct-C.  }
\setlength{\tabcolsep}{6pt}
\resizebox{\linewidth}{!}{
\begin{tabular}{lcc|cc}
    \toprule
        & \multicolumn{2}{c|}{UESTC-C} & \multicolumn{2}{c}{ HumanAct-C} \\
    Metrics &FID$\downarrow$ & \multicolumn{1}{c|}{Acc.$\uparrow$} 
    & FID$\downarrow$ & Acc.$\uparrow$ 
    \\
        \midrule
   using full class  & $102.48$    & $84.80$
   & $86.47$  & $63.6$    \\
w/o Gaussian &    $188.66$  & $59.90$    &$149.42$  &  $51.4$ \\
 w/o Mask & $294.87$  & $35.40$       & $251.85$  &   $31.6$   \\
Ours(w/o DR) & $102.48$    & $84.80$ & $86.41$ & $63.6$   \\
Ours(w/ DR)   & \textbf{90.13}    &  \textbf{87.20} & \textbf{83.43}  & \textbf{64.1}   \\
        \bottomrule
        
\end{tabular}
}
\label{tab:ablation}
\end{table}

\section{Experiments}

\subsection{Dataset \& Evaluation Metrics}

The HumanAct-C dataset contains a training set with 120 sequences for 12 sub-actions, and a testing set including 9 compositional categories with 900 compositional action sequences.  
As to the UESTC-C dataset, the testing set comprises 25 compositional categories with $625 $ compositional action sequences. The training set is composed of $ 50 $ sequences. Both the training and testing sets consist of the same 10 sub-actions. 
We employ the Frechet Inception Distance (FID) as our primary evaluation metrics.

\subsection{Implementation Details}
\label{sec:details}
\textbf{Network architecture:}
Our framework is comprised of a Transformer encoder and a decoder, as depicted in Figure~\ref{fig:framework}. The inputs to the Transformer encoder are the pseudo-compositional actions with mixed action embeddings, and 
the output represents the distribution parameters of the latent space.
The Transformer decoder utilizes a combination of the latent vector and the compositional label embedding for its key and value inputs, with 
a sequence of positional embeddings as the query. And the output is a sequence of 3D actions.

\textbf{Training details:}
The loss function of our model comprises three components: the reconstruction loss and the KL loss as per Eq.~\eqref{eq: cvae_loss}, and the Decoupling Refinement (DR) loss as defined by Eq.~\eqref{eq: mae_loss}. The balance rate between these components is set as $1: 1e-5: 1e-2$. 
We set the frame length to 60, with each frame containing $N=24$ joint points. 
Both the encoder and the decoder employ eight transformer layers. 
The AdamW is used as the optimizer with 1e-4 initial learning rate and no learning rate schedule is applied. We trained all models 1500 epochs with batch size 800. 


\subsection{Quantitative Evaluation}
We compare our method with three main baseline methods including Pre-train+Mask~\cite{actor}, Latent Disentanglement~\cite{arad2021compositional}, and the text-guided method MotionCLIP~\cite{motionclip}.  Pre-train+Mask is an intuitive compositional generation solution that directly composes two generated actions with attention masks. 
Table~\ref{tab:comparison} outlines our experimental results, along with those from baselines on the HumanAct-C and UESTC-C datasets. 
Following ACTOR~\cite{actor}, we report on FID and action recognition accuracy (Acc) as primary evaluation metrics, and also consider overall diversity and per-class diversity as auxiliary diversity metrics. Our method significantly outperforms baseline approaches and text-guided approaches on both datasets. 
The results underline the potential for our model to generate high-quality compositional motion sequences without 
extensive text annotations, leading the way for efficient compositional action generation.


\subsection{Ablation Studies}
We conduct ablation studies to investigate the effectiveness of different components. The performance results are summarized in Table~\ref{tab:ablation}.
It demonstrates the efficacy of the Decoupling Refinement in boosting the overall performance of our generation model. As revealed by the data in Table~\ref{tab:ablation}, the Decoupling Refinement consistently enhances our model's performance on both the HumanAct-C and UESTC-C datasets. 
Besides, it can be concluded that the pre-trained text-guided methods cannot be successfully  applied to our 
scenarios. As indicated in Table~\ref{tab:comparison}, MotionCLIP~\cite{motionclip} struggles to generate compositional actions. 
In our experiments, we utilize the MotionCLIP model pre-trained on AMASS~\cite{amass} (we're unable to train MotionCLIP using our dataset due to the absence of language annotations and compositional data). 
However, the results show that the model fails to generate satisfactory results.
In addition, it has been proven the energy-based attention mask plays a key role in coupling and decoupling.
In an experiment where the attention mask was set to $\bm{1}$, indicating equal scores for all body parts, the performance significantly declined.
This finding underscores the importance of the attention mask for learning disentangled representations.

\begin{figure}[t]
    \centering
\centerline{\includegraphics[width=\linewidth]{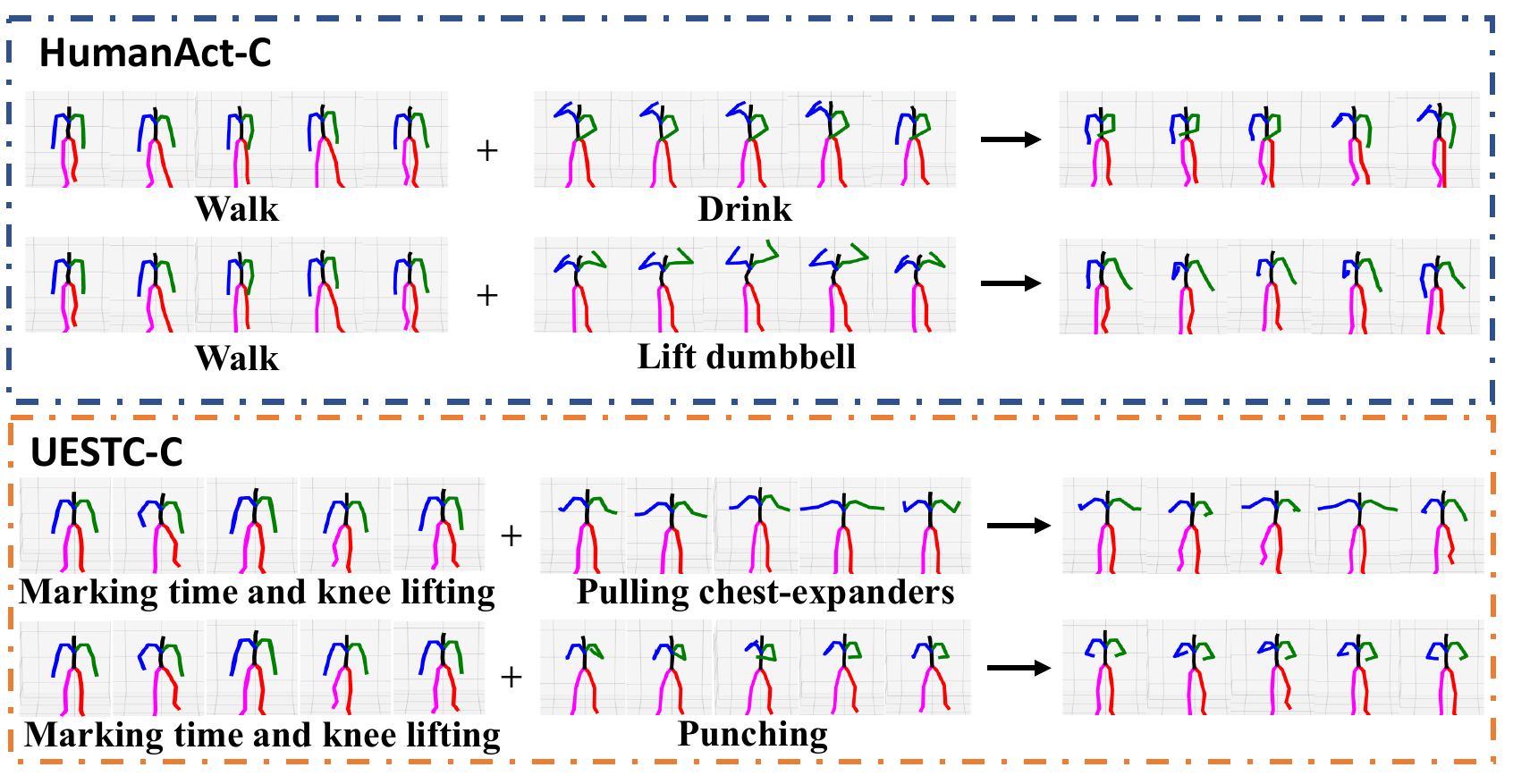}}
    \caption{ Compositional 3D motion generations with different categories on the HumanAct-C and UESTC-C datasets. 
  }
    \label{fig:skeleton-uestc-c}
\end{figure}


\begin{figure}[t]
    \centering
\centerline{\includegraphics[width=\linewidth]{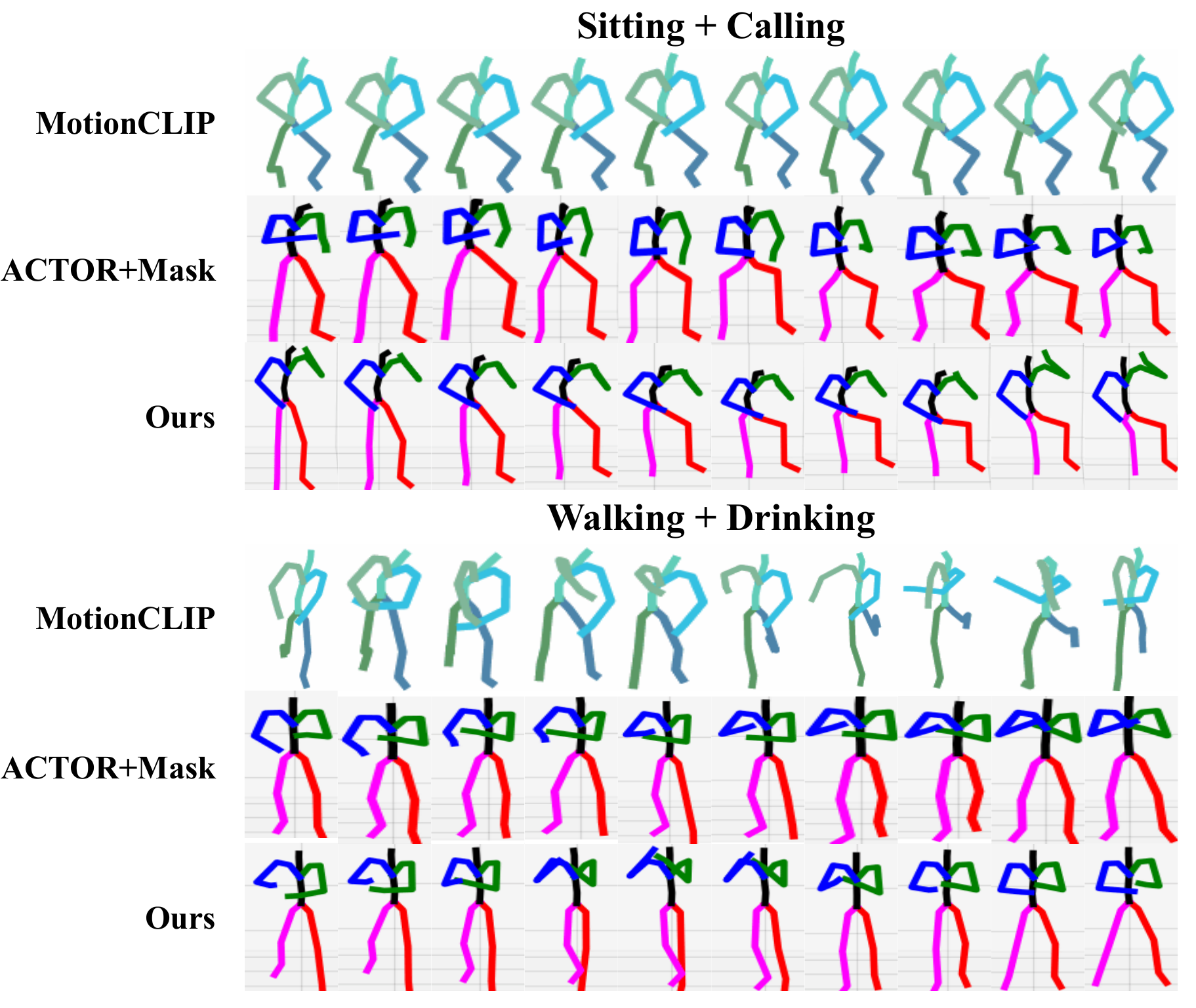}}
    \caption{ The qualitative comparisons  between our methods and other baseline methods. 
  }
    \label{fig:compare-2}
\end{figure}

\subsection{Qualitative Evaluation}
In this subsection, we showcase several visualization examples of the generated compositional 3D actions. Figure~\ref{fig:skeleton-uestc-c} displays results for four categories from the HumanAct-C and UESTC-C datasets. 
We see the generation of realistic compositional 3D motions. This validates our method's efficacy in learning the disentangled latent variable and its applicability to zero-shot compositional action generation. 
Moreover, Figure~\ref{fig:compare-2} compares the generations among the Pre-trained ACTOR~\cite{actor} + Mask, MotionCLIP~\cite{motionclip}, and ours.




\section{Conclusion}
In this paper, we present an innovative framework for generating compositional 3D action sequences, without the need of extensive annotations. 
We develop two novel datasets, HumanAct-C and UESTC-C, 
providing valuable resources for future research in this domain. Comprehensive evaluations of these datasets show our methodology's significant effectiveness in generating compositional actions, opening new avenues for further research in the field of 3D action generation.

\section*{\normalsize Acknowledgment}

This work was supported by the following grants: \\ Shenzhen Key Laboratory of Ubiquitous Data Enabling Grant No. ZDSYS20220527171406015, the UKRI grant of Turing AI Fellowship EP/W002981/1. We would also like to thank the Royal Academy of Engineering and FiveAI.





\bibliographystyle{IEEEbib}
\bibliography{cag}

\clearpage

\bigskip
%
\renewcommand{\thefigure}{A.\arabic{figure}} 
\setcounter{figure}{0} 
\renewcommand{\thetable}{A.\arabic{table}}
\setcounter{table}{0} 
\renewcommand{\thesection}{A.\arabic{section}}
\setcounter{section}{0} 

  \twocolumn[\textit{\large Appendix for}\\ \ \\
      {\large \bf ``Language-free Compositional Action Generation via Decoupling Refinement''}\
\vspace{.3cm}



{\large Appendix organization:}

\vspace{.1cm}

\DoToC
]

\section{More Related Work}
\textbf{Text-Conditioned Action Generation.} To augment the compositional ability in action generation, recent methods~\cite{motionclip,motiondiffuse,zhang2023remodiffuse} have proposed using neural language as the conditioning factor. This method capitalizes on the semantic structure inherent in language to empower composition ability. For instance, MotionCLIP~\cite{motionclip} employs CLIP to align the language semantic space with the visual motion space, and MotionDiffuse~\cite{motiondiffuse} uses language-annotated data to train a text-conditional diffusion model, enabling diverse and nuanced motion generation. However, these methods contrast with our research focus: we target a more challenging task that aims to generate simultaneous compositional actions with minimal annotation requirements.

\textbf{Mix-up.} Mix-up~\cite{zhang2018mixup} is a data augmentation technique that leverages convex linear interpolation to blend pairs of images and their corresponding targets. Its variations, such as CutMix~\cite{yun2019cutmix}, which overlays a cropped portion of an image onto another without changing the label, and Attentive CutMix~\cite{walawalkar2020attentive}, which employs an attention model to identify flexible cropping regions, have achieved considerable success. This paper introduces an innovative method utilizing an energy model to reasonably mix up active body parts to generate pseudo compositional actions.

\section{More Details of Method}
In this section, we outline our innovative framework designed for learning the compositional 3D action generation model without relying on previously collected composite data or annotated language descriptions. As illustrated in Figure~\ref{fig:framework}, the framework is founded on a tripartite process: \emph{Action Coupling}, \emph{Conditional Action Generation}, and \emph{Decoupling Refinement}. These integral components will be discussed in the subsequent subsections.

\subsection{Problem Definition}

The aim of language-free compositional 3D action generation is to cultivate a model capable of generating credible and diverse simultaneous compositions using two sub-action labels, even when limited annotations are given.
During the training phase, we work with paired training data composed of action sequences and their corresponding label information $\mathbf{x},\mathbf{y} \in \mathcal{X}\times\mathcal{Y} $. Here, $\mathbf{x}$ signifies the sub-action label, for instance, 'drinking', 'walking', or their embeddings, while $\mathbf{y} =\{\mathbf{y}_t\in \mathbb{R}^{N\times3}| t=1,2,\cdots, T \} $ represents the sequence of action poses. Specifically, each frame $\mathbf{y}_t$ comprises a pose that includes N points $\{\textbf{y}_{n,t}| n =1,2,...,N\} $, each of which is a 3D coordinate, with $T$ denoting the sequence length.
Beyond this sub-action sequence, there lies an unseen compositional action space $\mathcal{Y}\times\mathcal{Y}\rightarrow\mathcal{\tilde{Y}}$, wherein each composite action is constructed from two sub-actions. The aim then is to learn a generative model $g_{\psi}(\mathbf{\tilde{y}}|\mathbf{x}^i, \mathbf{x}^j,\mathbf{z})$ that produces a composite action $\mathbf{\tilde{y}} \in \mathcal{\tilde{Y}}$ given two different sub-action labels $\mathbf{x}^i, \mathbf{x}^j$ (here we use the superscript to represent different samples) and a noise vector $\mathbf{z}$.

\subsection{Action Coupling}
Unlike in text-guided generation tasks, the primary challenge in our scenario stems from the absence of composite actions in the training dataset. It is difficult and costly to capture the compositional actions, especially for simultaneous compositions. To solve this problem, we propose a method, Action Coupling, to synthesize the compositional action sequences from observed sub-actions. 

Given two label variables $\mathbf{x}^i, \mathbf{x}^j $ and corresponding sub-action sequences $\mathbf{y}^i, \mathbf{y}^j $, we couple two sub-action sequences to generate pseudo targets in the following procedure: 
 \begin{equation}
  \begin{aligned}
\label{eq: mix_up} \tilde{\mathbf{x}} &= \lambda f_x^i(\mathbf{x}^i) + (1-\lambda) f_x^j(\mathbf{x}^j), \\
\tilde{\mathbf{y}} &= \lambda f_y^j(\mathbf{y}^j) + (1-\lambda) f_y^j(\mathbf{y}^j),
  \end{aligned}
\end{equation}
where the coefficient $\lambda \in [0,1]$ denotes the mixing rate between two source sub-actions, and functions $f_x, f_y$ represent the coupling strategy. In the following, we will discuss how to design the distribution of $\lambda$ and the implementation of functions $f_x, f_y$.

\textbf{Mixing rate distribution:}
In the context of Mix-up methods for data augmentation~\cite{zhang2018mixup}, the parameter $\lambda$ is sampled from $\lambda \sim Beta(\alpha,\alpha)$, where $\alpha \in (0,\inf)$ is a hyper-parameter that regulates the intensity of interpolation. This Beta distribution pushes the $\lambda$ close to $0$ or $1$ and away from $0.5$. It is reasonable, as overly mixed images (e.g., where $\lambda=0.5$) may dilute discriminative capacity, thereby contradicting the objective of data augmentation, which is to enhance classification.
Conversely, our methodology is centered around generating composite actions that preserve as much information from both sub-actions as possible. In essence, we promote the deep integration of two actions, ideally where $\lambda =0.5$. As such, we employ $\lambda \sim \mathcal{N}(0.5,\delta)$, where $\delta$ denotes the variance of the Gaussian distribution.

\textbf{Preliminary insights:} Prior to exploring the mechanics of $f_x, f_y$, we aim to convey key observations about human actions that serve as the motivation for our following designs. Firstly, it is important to recognize that \emph{a single body part is incapable of executing two distinct actions simultaneously}. For instance, it is plausible to perform actions like "drinking while walking", but attempting to do "calling and singing" concurrently doesn't make practical sense. This realization implies that simultaneous compositional actions generally involve two different body parts carrying out different sub-actions. This insight motivates our strategy of crafting two attention masks that focus on disparate body parts during the sub-action coupling process.
Secondly, we found that \emph{body parts with the highest motion energy tend to be crucial in defining a particular action}. Essentially, significant motion is almost always associated with a body part that plays a key role in a certain sub-action, such as the ``leg'' part in the sub-action ``walking''.  This insight drives us to calculate motion energy to identify the active body parts.

\textbf{Energy-based attention masks:} Informed by the insights above, we formulate the subsequent energy model to compute the motion energy for each individual body part:
 \begin{equation}
  \begin{aligned}
\label{eq: motion energy} 
\mathbf{E}_{n} = \frac{1}{|\Omega|(T-1)}\sum_{\mathbf{y}_n\in \Omega }\sum_{t=2}^T ||\mathbf{y}_{n,t} - \mathbf{y}_{n,t-1}||^2,
  \end{aligned}
\end{equation}
where $\Omega$ denotes a body part, such as ``leg'' or ``arm'', containing a set of joint points $\{\textbf{y}_{n,t}| n \in \Omega \}$, and $n,t$ denotes the $n$th joint point in the $t$th frame. Please note that for the joint points in the same body part, the energy values are equal.
Then, we can consequently define the attention value as the energy value, e.g., if the motion energy of the ``arm'' part is higher, then all the points in the ``arm'' part will be assigned to high attention scores. 
Finally, we couple two sub-actions with the obtained attention mask $\mathbf{E}_{n}$ and implement the functions $f_y$ in the second line in Eq~\eqref{eq: mix_up} as
 \begin{equation}
  \begin{aligned}
\label{eq: mix_up2} 
\tilde{\mathbf{y}}_n &= \frac{\lambda\textbf{E}^{i}_{n}\mathbf{y}^{i}_{n}}{\lambda\textbf{E}^{i}_{n} + (1-\lambda) \textbf{E}^{j}_{n}} + \frac{(1-\lambda) \textbf{E}^{j}_{n}\mathbf{y}^{j}_n}{\lambda\textbf{E}^{i}_{n} + (1-\lambda) \textbf{E}^{j}_{n}}.
  \end{aligned}
\end{equation}
While $f_x$ is implemented by the identity function, i.e., $ \tilde{\mathbf{x}} = \lambda \mathbf{x}^i + (1-\lambda) \mathbf{x}^j$.

\subsection{Conditional Action Generation} 
Given the generated pseudo compositional training data $\tilde{\mathbf{y}}$ and $\tilde{\mathbf{x}}$, we formulate the generative process as $p(\tilde{\mathbf{y}}|\tilde{\mathbf{x}}) $:
 \begin{equation}
  \begin{aligned}
\label{eq: prediction_process} 
p(\tilde{\mathbf{y}}|\tilde{\mathbf{x}}) = \int g_{\psi}(\tilde{\mathbf{y}}|\tilde{\mathbf{x}},\mathbf{z})p_{\theta}(\mathbf{z}|\tilde{\mathbf{x}})d{\mathbf{z}}.
  \end{aligned}
\end{equation}
This generative process can be learned with Conditional VAE (CVAE)~\cite{cvae}, where a variational neural network $q_{\phi}(\mathbf{z}|\tilde{\mathbf{x}},\tilde{\mathbf{y}}) $
is introduced to optimize the objective above. The optimization follows the standard variational lower bound:
\begin{equation}
  \begin{aligned}
\label{eq: cvae_loss} \mathcal{L}_{CVAE}(\psi,\phi,\theta) =   \mathcal{L}_{KL}\big(q_{\phi}(\mathbf{z}|\tilde{\mathbf{x}},\tilde{\mathbf{y}})||p_{\theta}(\mathbf{z}|\tilde{\mathbf{x}})\big) \\
-\lambda \mathbb{E}_{q_{\phi}(\mathbf{z}|\tilde{\mathbf{x}},\tilde{\mathbf{y}})}[\log g_{\psi}(\tilde{\mathbf{y}}|\tilde{\mathbf{x}},\mathbf{z})].  
  \end{aligned}
\end{equation}
The first term reduces the KL divergence between $q_{\phi}(\mathbf{z}|\tilde{\mathbf{x}},\tilde{\mathbf{y}}) $ and  $p_{\theta}(\mathbf{z}|\tilde{\mathbf{x}})$, which encourages the consistency of latent variables generated from condition and data. The second term is a reconstruction objective that encourages the model to generate more realistic action poses.

\subsection{ Decoupling Refinement}
Considering the potential compromise in the quality of generated sequences due to suboptimal coupling, we introduce \emph{Decoupling Refinement}. This strategic implementation serves as a constraint to further enhance the quality of the generated actions. Central to its design is the application of knowledge from self-supervised pre-trained models, to ensure the semantic consistency between the sub-actions and the generated compositional actions. 

\textbf{3D to 2D rendering:} In this study, we utilize the pre-trained MAE~\cite{mae} as our self-supervised model. Opting for MAE over alternatives like CLIP is informed by two primary considerations: 1) Our scenario lacks textual annotations, and 2) The MAE model is widely used with its robust image inpainting abilities. However, there is a gap between the 3D action sequence and the images (used for MAE). To bridge this gap, we utilize the Skinned Multi-Person Linear (SMPL) body model~\cite{loper2015smpl}.  Both the original sub-actions and the generated sequences are processed through the SMPL model, producing the dense point cloud representing human actions with the corresponding body surfaces. These 3D models are then projected onto a 2D plane using a rendering function. This process can be formulated as  $f_{R} : \{\textbf{y}^i, \textbf{y}^j, \hat{\textbf{y}} \in \mathbb{R}^{3\times N}\} \rightarrow \{\textbf{v}^i, \textbf{v}^j, \hat{\textbf{v}} \in \mathbb{R}^{3\times H \times W}\}$, where $\textbf{v}$ denotes the rendering images with height $H$ and width $W$.
To enhance the readability of the resultant 2D images, we standardize the 3D model to ensure rendering from a frontal perspective.

\textbf{Decoupling:} Drawing inspiration from the coupling process, we introduce a reversal of this process, referred to as decoupling. This step allows us to deconstruct the rending images of compositional actions back into their sub-action components and further facilitate refining the generation.

Intuitively, for a compositional action like ``drinking and walking'', our goal is to preserve the active parts associated with each sub-action. For example, we aim to retain the ``arms'' when considering the ``drinking'' sub-action, and the ``legs'' for the ``walking'' sub-action. Technically, we propose a 3D-2D attention projection to compute the decoupling attention map using the motion energy $\textbf{E}_n$. 
By utilizing the SMPL projection and the rendering function $f_R$,  we first transform each 3D joint point $\textbf{y}^i_n $ into 2D coordinates $\textbf{pix}^i_n $. 
The pixels in proximity to active joint points should be prioritized to focus. Hence, we establish a decay rate that is inversely proportional to the square of the pixel distance. We then compute the value for each pixel in the 2D attention map $\mathcal{A}$ by aggregating these decayed attention scores from all $N$ joint points:
 \begin{equation}
  \begin{aligned}
\label{eq: attention} 
\mathcal{A}^i(\textbf{pix}) = \sum_{n=1}^{N} \frac{\textbf{E}^i_n}{||\textbf{pix}-\textbf{pix}^i_n||^2},
  \end{aligned}
\end{equation}
where $\textbf{pix}$ denotes each pixel of the rendering image $\textbf{v}$. Lastly, we divide the image into segments of $16\times16$ regions, following the MAE model's configuration. We then compute the average attention value for each region and retain a subset of regions (e.g., 1/3) with high attention and masked the others. With this process, we obtain the decoupled images $\hat{\textbf{v}}^i=\hat{\textbf{v}} \circ \mathcal{A}^i $ 
and $\hat{\textbf{v}}^j=\hat{\textbf{v}} \circ \mathcal{A}^j$ from the compositional image.

\textbf{Constraints:} The masked images are input into a pre-trained MAE model for the inpainting process. Despite the input image being part of a generated composite action, we encourage the model to reconstruct the original sub-actions, denoted as $\hat{\textbf{v}}_{mae}^i,\hat{\textbf{v}}_{mae}^j = f_{mae}(\hat{\textbf{v}}^i,\hat{\textbf{v}}^j )$, where $f_{mae}$ denotes the MAE model. It indicates that the attributes of the sub-actions are preserved during the compositional generation process and can be readily disentangled. For all data pairs $\textbf{y}^i,\textbf{y}^j$, the loss function for the decoupling refinement process can be expressed as follows:
 \begin{equation}
  \begin{aligned}
\label{eq: mae_loss} \mathcal{L}_{DR}= \sum_{i,j} (||\hat{\textbf{v}}_{mae}^i-\textbf{v}^i||^2 +||\hat{\textbf{v}}_{mae}^j-\textbf{v}^j||^2) .
  \end{aligned}
\end{equation}
Finally, the CVAE loss $\mathcal{L}_{CVAE}$ and the Decoupling Refinement loss $\mathcal{L}_{DR}$ are integrated to train our model.

\section{More Experiments}

To evaluate the performance of our proposed framework, we have created two new datasets and corresponding evaluation metrics. We then employ these metrics to compare our method against other compositional generation methods, including those utilizing textual annotations. An ablation study is also conducted to examine the impact of individual components on the overall performance. Additionally, we present a qualitative analysis for a deeper understanding of our method.

\begin{figure*}[t]
    \centering
\centerline{\includegraphics[width=\linewidth]{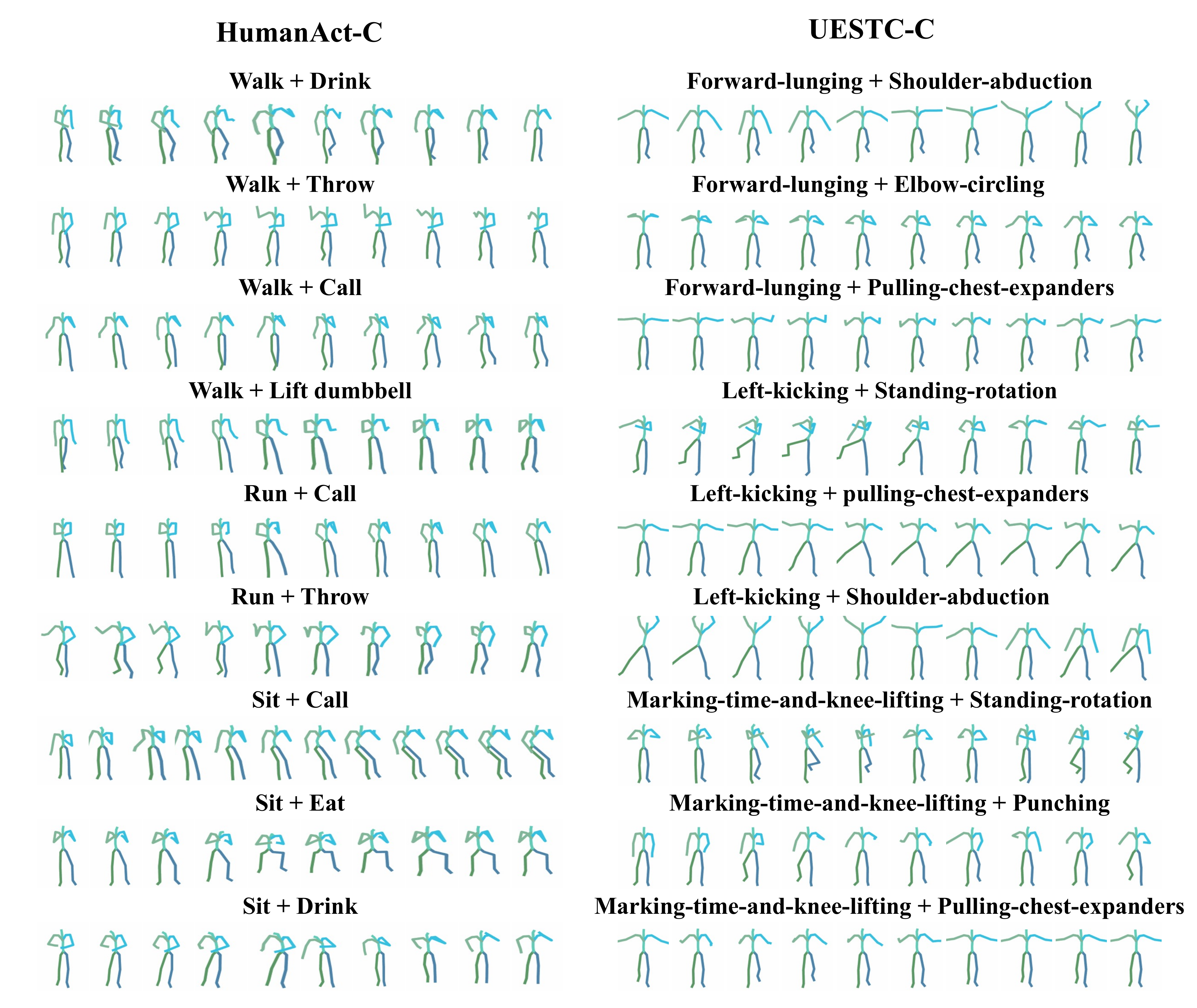}}
    \caption{ The visualization of examples of our dataset HumanAct-C and UESTC-C.
  }
    \label{fig:dataset}
\end{figure*}

\subsection{Dataset \& Evaluation Metrics}
Given the lack of available datasets that incorporate both sub-actions and compositional actions, we have constructed two novel datasets: HumanAct-C and UESTC-C. These are founded on the HumanAct12~\cite{guo2020action2motion} and UESTC~\cite{ji2018large} datasets, respectively.



Given the lack of available datasets that incorporate both sub-actions and compositional actions, we have constructed two novel datasets: HumanAct-C and UESTC-C. These are founded on the HumanAct12~\cite{guo2020action2motion} and UESTC~\cite{ji2018large} datasets, respectively.

\textbf{HumanAct-C dataset:} 
The original HumanAct12 dataset includes 12 action categories, annotated with joint coordinates and SMPL pose parameters processed by~\cite{actor}. We utilized these categories as sub-actions, from which we can generate compositional action categories by combining each pair of actions to build the HumanAct-C dataset. HumanAct-C is organized into a training set of original sub-actions and a testing set of compositional actions. The training set is derived from a subset of the original HumanAct12 training set. As for the testing set, we synthesize compositional actions using testing samples from a pair of action sequences with differing labels. These sequences are fused using motion energy-based attention masks, and manually curated to remove any irrational samples and unfeasible combination categories, such as ``drinking while phoning''. In the end, we curate a training set of 120 sequences for 12 sub-actions, and a testing set featuring 9 compositional categories with a total of 900 compositional action sequences. 
These 9 compositional categories are  based on 7 commonly seen sub-actions: "walk+drink", "walk+call", "walk+throw", "walk+ lift dumbbell", "run+call", "run+throw", "sit+call", "sit+eat", and "sit+drink". 
Please note that the sub-action categories in training and testing sets may not be strictly aligned. For example, there are 12 sub-actions for training, and 7 of them are used to build compositional actions. The examples of these 9 compositional actions are visualized in Figure~\ref{fig:dataset}.

\textbf{UESTC-C dataset:} The UESTC dataset consists of 25K sequences across 40 action categories, which also include both joint coordinates annotations and SMPL pose parameters. Through the composition, we generated 780 potential categories and select 25 feasible categories for evaluation, with these categories being based on 10 sub-actions.
Finally, we generate the testing set comprising 25 compositional categories with a total of $25\times 25 = 625 $ compositional action sequences. The training set is composed of $ 10 \times 5 =50 $ sequences. These compositional action categories are summarized below: "marking-time-and-knee-lifting + pulling-chest-expanders", "marking
time-and-knee-lifting + punching", "marking-time-and-knee-lifting + elbow-circling", "marking time-and-knee-lifting + shoulder-abduction", "marking-time-and-knee-lifting + standing-rotation", "punching-and-knee-lifting + pulling-chest-expanders", "punching-and-knee-lifting+punching", "punching-and-knee-lifting + elbow-circling", "punching-and-knee-lifting + shoulder-abduction", "punching-and-knee-lifting + standing-rotation", "left-lunging+pulling-chest-expanders", "left lunging + punching", "left-lunging + elbow-circling", "left-lunging + shoulder-abduction", "left lunging + standing-rotation", "left-kicking + pulling-chest-expanders", "left-kicking + punching",
"left-kicking + elbow-circling", "left-kicking + shoulder-abduction", "left-kicking + standing-rotation", "forward-lunging + pulling-chest-expanders", "forward-lunging + punching", "forward-lunging + elbow-circling", ""forward-lunging + shoulder-abduction", ""forward-lunging + standing-rotation".
We also provide the visualization of the synthetic compositional data in Figure~\ref{fig:dataset}.

\textbf{Evaluation metrics:} 
To quantitatively assess the capability of compositional action generation, we introduce a novel metric derived from the Frechet Inception Distance (FID), which we term FID-C. This involves training a compositional action recognition model STGCN~\cite{yu2017spatio}, on the testing data set. Subsequently, FID-C is computed as 
 \begin{equation}
  \begin{aligned}
\label{eq: fid} 
\text{FID-C} = ||\mu_t -\mu_g ||^2 + Tr(\Sigma_t+\Sigma_g -2(\Sigma_t\Sigma_g)^{1/2})
  \end{aligned}
\end{equation}
where $\mu_t$ and $\Sigma_t$ represent the mean and variance of the STGCN features for the testing data, while $\mu_g$ and $\Sigma_g$ correspond to those for the data generated by our trained model.

\subsection{Implementation Details}
\label{sec:details}

\textbf{Network Architecture:}
In our approach, we use ACTOR~\cite{actor} as the baseline, comprising a Transformer encoder and decoder, depicted in Figure~\ref{fig:framework}. The inputs to the Transformer encoder are the pseudo-compositional actions and mixed action embeddings, which are derived from our action coupling technique. These inputs, composed of tokenized label embeddings and frame-by-frame compositional actions, are concatenated together and appended with position embeddings to create the final inputs. The output of the Transformer encoder represents the distribution parameters of the latent space, enabling the sampling of a latent vector for action generation.

The Transformer decoder utilizes a combination of the latent vector and the compositional label embedding for its key and value inputs, while a sequence of positional embeddings is used as the query. The decoder's output is a sequence of 3D actions, matching the size of the input pseudo-actions. To further refine these 3D actions and acquire the corresponding body surfaces, we employ the SMPL model.

To enhance the model's performance, we incorporated elements from the ACTOR framework as referenced in the literature. In line with ACTOR's recommendations, we applied the Gaussian Linear Error Units (GELU) activation function within the Transformer architecture. This function is known for improving the learning dynamics and performance of the Transformer.

\textbf{Hyper-parameters:}
The final loss function of our model comprises three components: the reconstruction loss and the KL loss as per Eq.~\eqref{eq: cvae_loss}, and the Decoupling Refinement (DR) loss as defined by Eq.~\eqref{eq: mae_loss}. The balance rate between these components is set as $1: 1e-5: 1e-2$. To ensure an efficient training, we follow ACTOR's~\cite{actor} example by setting the frame length to 64, with each frame containing $N=24$ joint points. These points are projected into a continuous 6D rotation representation space $\mathbb{R}^{24\times6}$.
During the action coupling process, the standard deviation $\sigma$ of the Gaussian distribution is set to 0.1. In the presented experiments, an 8-layer Encoder and Decoder were utilized within the Transformer model. Multi-head attention was configured with 4 heads, and a dropout rate of 0.1 was applied to prevent overfitting and enhance the generalization ability of the model. The input embedding dimension was set at 256, and the intermediate feedforward network had a dimension of 1024. Regarding the MAE model, we utilize a larger vision that also incorporates a GAN loss to enhance the in-painting capabilities.

\textbf{Training details:}
The AdamW~\cite{adamw} is used as the optimizer with 1e-4 initial learning rate and no learning rate schedule is applied. We trained all models 1500 epochs with batch size 800. All experiments are conducted with Pytorch~\cite{paszke2019pytorch} 1.8.1 and trained on 8 NVIDIA A100 GPUs. It took about 2 days and 1 day respectively for the model training on the HumanAct-C and UESTC-C datasets.

\subsection{Quantitative Evaluation}
We compare our method with three main baseline methods including Pre-train+Mask (ACTOR)\cite{actor}, Latent Disentanglement~\cite{arad2021compositional}, and the text-guided method MotionCLIP~\cite{motionclip}.  Pre-train+Mask is an intuitive solution for the compositional generation to directly compose two generated actions with attention masks. Latent disentanglement learns the latent variables for more diverse generations. However, they didn't involve the action coupling process to produce more diverse training data and the decoupling refinement process to improve the action quality. The text-guided methods have been discussed in Section A.1. 


\textbf{ACTOR+Mask:} The ACTOR method, originally designed for conventional 3D motion generation, lacks the capability for compositional generalization. To address this, we adopt an intuitive solution based on attention masking. As a simple baseline, we initially train an ACTOR for single actions, subsequently extracting the attention maps to compose two generated actions. For instance, given a composite label such as "drinking while walking", we generate two separate action sequences — "drinking" and "walking" — using the pre-trained ACTOR model. We then learn two attention masks to identify the key body parts involved in each action and generate the final composite actions.

\textbf{Latent Disentanglement:} To learn the disentangled latent variable, we require data comprising multiple factors. However, in our setup, we only have sub-action training data, each of which contains a single factor per class. As a result, it directly fuses two sub-actions by fixing $\lambda$ to obtain the data, and learns the  disentangled latent variable.  This baseline differs from our method in two primary ways.
However, they didn't involve the action coupling process to produce more diverse training data and the decoupling refinement process to improve the action quality.

\textbf{MotionCLIP:} MotionCLIP~\cite{motionclip} is a 3D human motion auto-encoder that aligns its latent space with the CLIP model. This alignment enables it to offer capabilities like out-of-domain actions, disentangled editing, and abstract language specification. Additionally, it uses a transformer-based auto-encoder to reconstruct motion while aligning with its text label's position in the CLIP space, enhancing its text-to-motion capabilities. In our setting, we use the pre-trained MotionCLIP on the AMASS\cite{amass} dataset, since our dataset didn't involve the compositional actions and well-annotated texts. To use the pre-trained model, we combine the labels of two sub-actions with ``while'' to highlight that this is a simultaneous composition, such as ``sitting while drinking''.

\textbf{Comparison and analysis:} Table~\ref{tab:comparison} outlines our experimental results, along with those from other methodologies on the HumanAct-C and UESTC-C datasets. Mirroring the metrics used in Action2motion~\cite{guo2020action2motion} and ACTOR~\cite{actor}, we report on FID-C and action recognition accuracy (Acc) as primary evaluation metrics, and also consider overall diversity and per-class diversity as auxiliary diversity metrics. Our method significantly outperforms baseline approaches on both datasets. For instance, compared to the baseline method Latent Disentanglement~\cite{arad2021compositional}, our technique shows an improvement of 39.65 in FID-C and 11.9 in accuracy on HumanAct-C, and an enhancement of 31.14 and 4.0 on FID-C and accuracy, respectively, on UESTC-C. Furthermore, our method surpasses the performance of text-guided approaches.
This result underlines the potential for our model to generate high-quality compositional action sequences without reliance on costly and extensive text annotations, leading the way for efficient compositional action generation.

\begin{figure}[t]
    \centering
\centerline{\includegraphics[width=\linewidth]{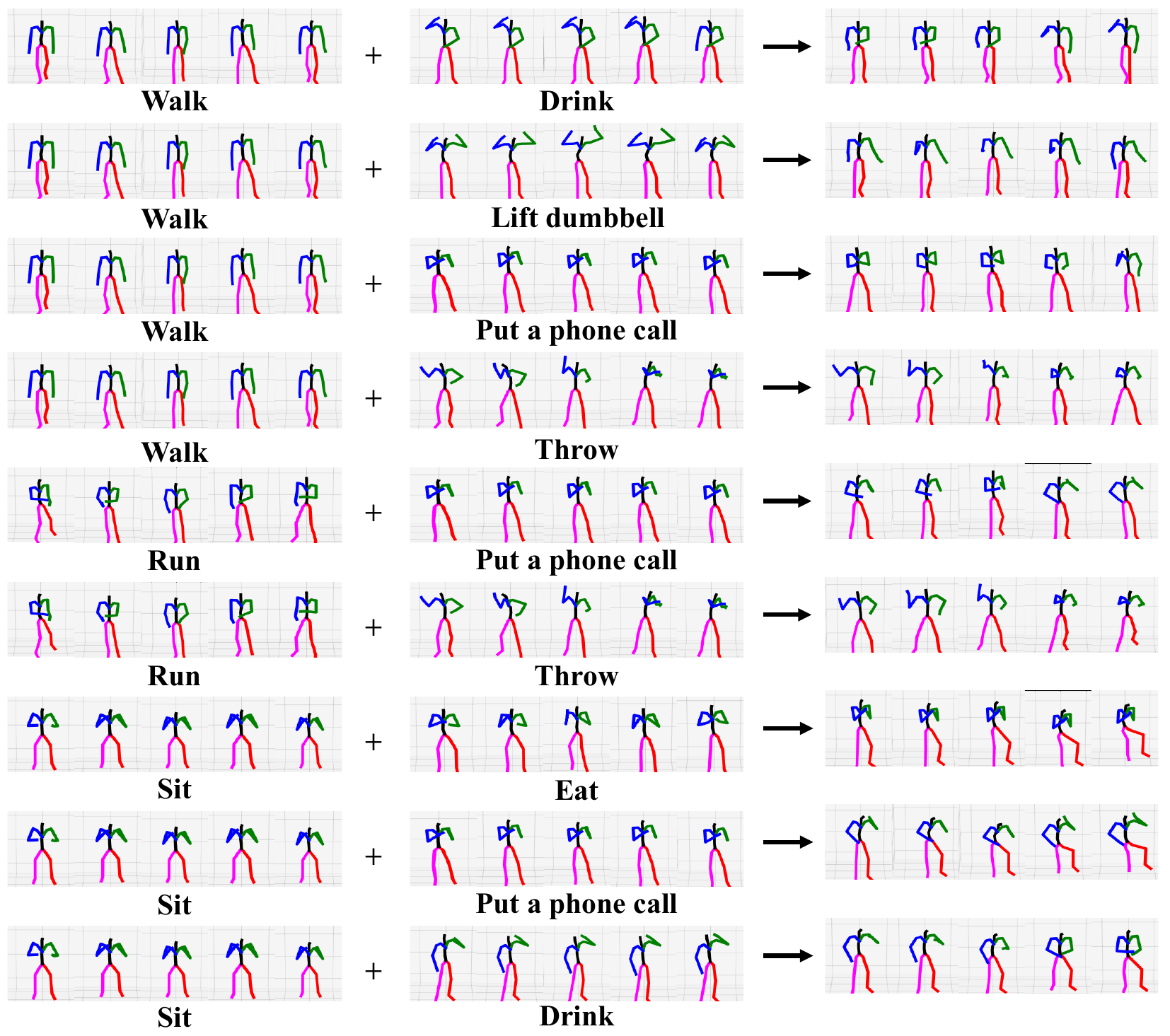}}
    \caption{ Compositional 3D motion generations with 9 categories on HumanAct-C. We provide the visualizations of a pair of raw sub-actions and the compositional actions generated 
 by our method. Zoom in for details.
  }
    \label{fig:skeleton}
\end{figure}

\subsection{Ablation Studies}
\begin{table*}
\centering
\caption{ Ablation studies on UESTC-C and HumanAct-C.  }
\setlength{\tabcolsep}{6pt}
\resizebox{.99\linewidth}{!}{
\begin{tabular}{lccrc|cccc}
    \toprule
        & \multicolumn{4}{c|}{UESTC-C} & \multicolumn{4}{c}{ HumanAct-C} \\
    Metrics 
    & FID-C$\downarrow$ & \multicolumn{1}{c}{Acc.$\uparrow$} & Div.$\rightarrow$ & Multimod.$\rightarrow$ & FID-C$\downarrow$ & Acc.$\uparrow$ & Div.$\rightarrow$ & Multimod.$\rightarrow$ \\
        \midrule
          using full class 
          &\textbf{$102.48^{\pm22.12}$} & \textbf{$84.80^{\pm1.59 }$} & \textbf{$33.25^{\pm0.61}$} & \textbf{$10.56^{\pm0.12}$}  & \textbf{$86.47^{\pm9.98}$ } & \textbf{$63.6^{\pm2.62}$} & \textbf{$21.31^{\pm0.57}$} & \textbf{$14.04^{\pm0.32}$}
          \\
    w/o Gaussian & \textbf{$188.66^{\pm33.41}$} & \textbf{$59.90^{\pm6.29}$ } & \textbf{$29.60 ^{\pm0.35}$} & \textbf{$ 13.12^{\pm0.11}$}
   & \textbf{$149.42^{\pm24.48}$} & \textbf{$51.4^{\pm4.54}$ } & \textbf{$21.38^{\pm0.50}$} & \textbf{$14.77^{\pm0.31}$}
      \\
   w/o Mask & \textbf{$294.87^{\pm71.4}$ } & \textbf{$35.40^{\pm4.30}$ } & \textbf{$25.77^{\pm0.41}$} & \textbf{$12.86^{\pm0.27}$}  &
   \textbf{$251.85^{\pm37.47}$}  & \textbf{$31.6^{\pm2.54}$ } & \textbf{$18.39^{\pm0.56}$} & \textbf{$13.92^{\pm0.66}$} \\

    Ours(w/o DR)    &\textbf{$102.48^{\pm22.12}$} & \textbf{$84.80^{\pm1.59 }$} & \textbf{$33.25^{\pm0.61}$} & \textbf{$10.56^{\pm0.12}$}  &\textbf{$86.41^{\pm19.8}$} & \textbf{$63.6^{\pm5.26}$} & \textbf{$21.11^{\pm0.58}$} & \textbf{$14.29^{\pm0.13}$} \\
Ours(w/ DR) & \textbf{$90.13^{\pm12.58}$ } & \textbf{$87.20^{\pm1.69}$} & \textbf{$33.06^{\pm0.81}$} & \textbf{$10.12^{\pm0.06}$} 
& \textbf{$83.43^{\pm13.01}$ } & \textbf{$64.1^{\pm5.41}$} & \textbf{$21.36^{\pm0.39}$} & \textbf{$14.01^{\pm0.38}$}\\
        \bottomrule
\end{tabular}

}
\label{tab:comparison_supp}
\end{table*}

We also conduct ablation studies to investigate the effectiveness of different components. In Table~\ref{tab:comparison_supp}, we display the comprehensive outcomes of our ablation studies, which further include a meticulous examination of our model's performance based on the diversity metric. The detailed results attest to the efficacy of our method, which serves as the evidence to answer the following questions:

\textbf{Q: How effective is the proposed decoupling refinement? } \textbf{A: It has proven highly effective.} As revealed by the data in Table~\ref{tab:ablation}, the inclusion of Decoupling Refinement consistently enhances our model's performance on both the HumanAct-C and UESTC-C datasets. Notably, the implementation of this technique brings about an improvement of 3.0 in FID-C and 0.5 in accuracy on the HumanAct-C dataset, while on the UESTC-C dataset, it increases FID-C by 12.4 and accuracy by 2.4. This clearly demonstrates the efficacy of the Decoupling Refinement method in boosting the overall performance of our action generation model.

\textbf{Q: Can the pre-trained text-guided methods be successfully applied to our scenarios? } \textbf{A: No.} As indicated in Table~\ref{tab:comparison}, MotionCLIP~\cite{motionclip} struggles to generate compositional actions when its training data lacks corresponding compositional actions. Specifically, in our experiments, we utilize the MotionCLIP model pre-trained on AMASS~\cite{amass} (we're unable to train MotionCLIP using our dataset due to the absence of language annotations and compositional data). During inference, we use the phrase "action1 while action2" as the text prompt for simultaneous compositional actions for MotionCLIP. However, the results show that the model fails to generate satisfactory results when the test data extends beyond the scope of the training text. This is further evidenced by the visualizations provided in Figure~\ref{fig:compare}.


\textbf{Q: What is the effect of the energy-based attention mask? } \textbf{A: It is the key to coupling and decoupling.} In an experimental setting where the attention mask was set to $\bm{1}$, indicating equal scores for all body parts, the performance significantly declined, with an FID score of 251.85 and an accuracy rate of 31.6. This was notably worse than baseline methods. This finding underscores the importance of the attention mask for learning disentangled representations, as crucial attributes of sub-actions tend to interact with each other. Additionally, a visual comparison, which further illustrates the impact of our energy-based attention mask, has been provided in Figure~\ref{fig:compare_energy}.

\textbf{Q: What is the effect of using a uniform distribution instead of a Gaussian distribution for coupling?} \textbf{A: Substantial performance decline.} Switching the distribution of $\lambda$ from Gaussian $\lambda \sim \mathcal{N}(0.5,\delta)$ to uniform leads to a significant drop in performance, with FID soaring from 86.41 to 149.42, and accuracy descending from 63.6 to 51.4. This is because our objective is to generate compositional actions where $\lambda$ is approximately $0.5$. Rates near 0 or 1 do not provide adequate supervision for learning the characteristics of compositional actions.

\textbf{Q: Does employing more compositional categories during training enhance performance?} \textbf{A: Not necessarily.}
We compare the performance when using a subset of categories for training versus using all compositions (both reasonable and unreasonable actions). For example, for HumanAct-C, the subset consists of 9 categories selected for testing and all compositions have 66 categories.  We find that both scenarios yielded comparable performance. This suggests that incorporating additional categories, even if they lack semantic meaning, doesn't negatively impact the results. Moreover, it demonstrates that our method does not require prior knowledge of which compositional categories are meaningful. Training the model exclusively with sub-action data is sufficient, making it highly suitable for real-world applications where unseen compositional scenarios may often occur.

\begin{figure}[t]
    \centering
    \centerline{\includegraphics[width=1.0\linewidth]{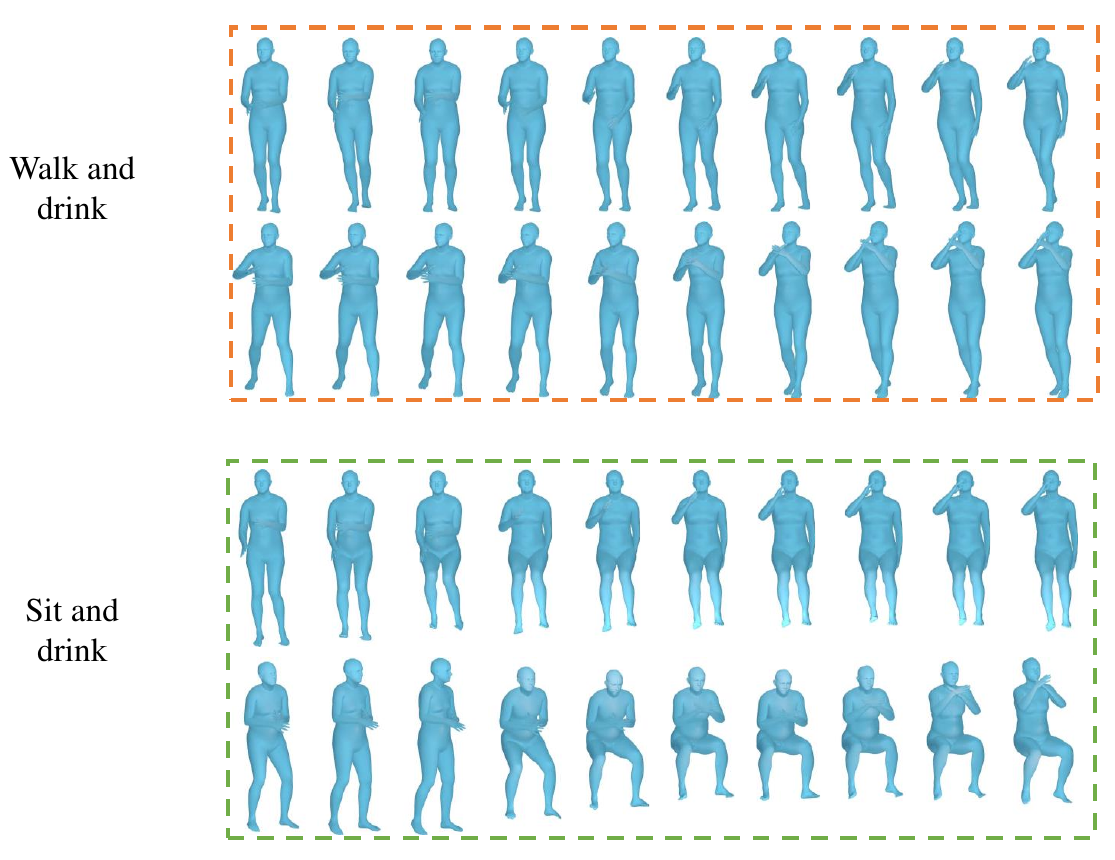}}
    \caption{ Visualization using SMPL with different sampled sequences of the same compositional categories. 
  }
    \label{fig:smpl}
\end{figure}

\subsection{Qualitative Evaluation}
\begin{figure*}[t]
    \centering
\centerline{\includegraphics[width=1.0\linewidth]{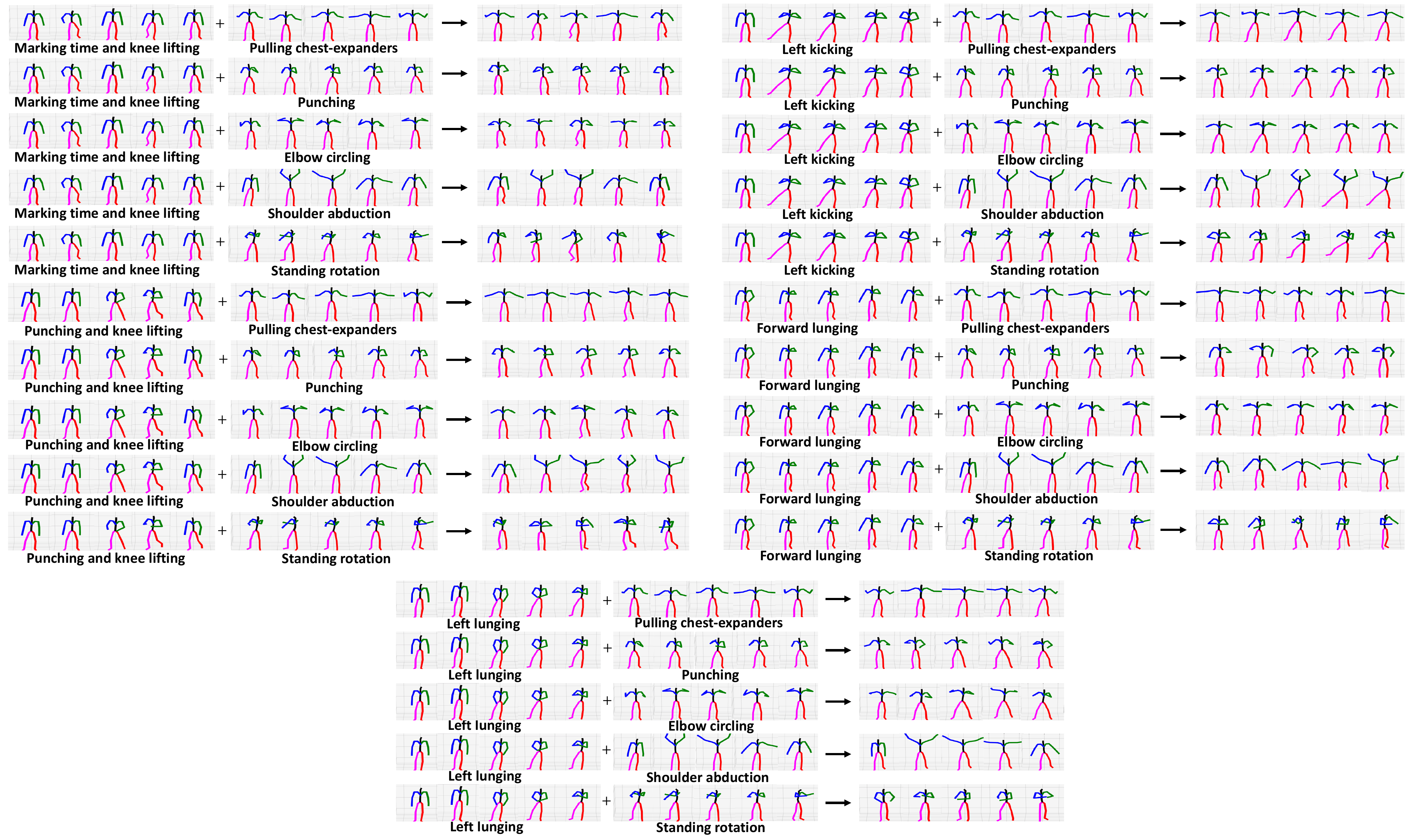}}
    \caption{ Compositional 3D motion generations with different categories. We provide the visualizations of a pair of raw sub-actions and our generated compositional actions on all 25 compositional categories. Zoom in for the details.
  }
    \label{fig:skeleton-uestc2}
\end{figure*}

In this subsection, we showcase several visualization examples of the generated compositional 3D actions. Figure~\ref{fig:skeleton} displays results for nine categories from the HumanAct-C dataset. 
For instance, observing the "drink and walk" example, we see the generation of realistic compositional 3D motions. This validates our method's efficacy in learning the disentangled latent variable and its applicability to zero-shot compositional action generation. 
Moreover, Figure~\ref{fig:smpl} presents multiple sequences for a single compositional category using SMPL~\cite{loper2015smpl} to highlight the diversity of our generations, where two sequences each for "walk and drink" and "sit and drink" are displayed.
Additionally, We provide comprehensive visualizations of our model’s performance
on the UESTC-C dataset in Figure~\ref{fig:skeleton-uestc2}. The dataset comprises all 25 categories of actions. 

\begin{figure}[t]
    \centering
\centerline{\includegraphics[width=\linewidth]{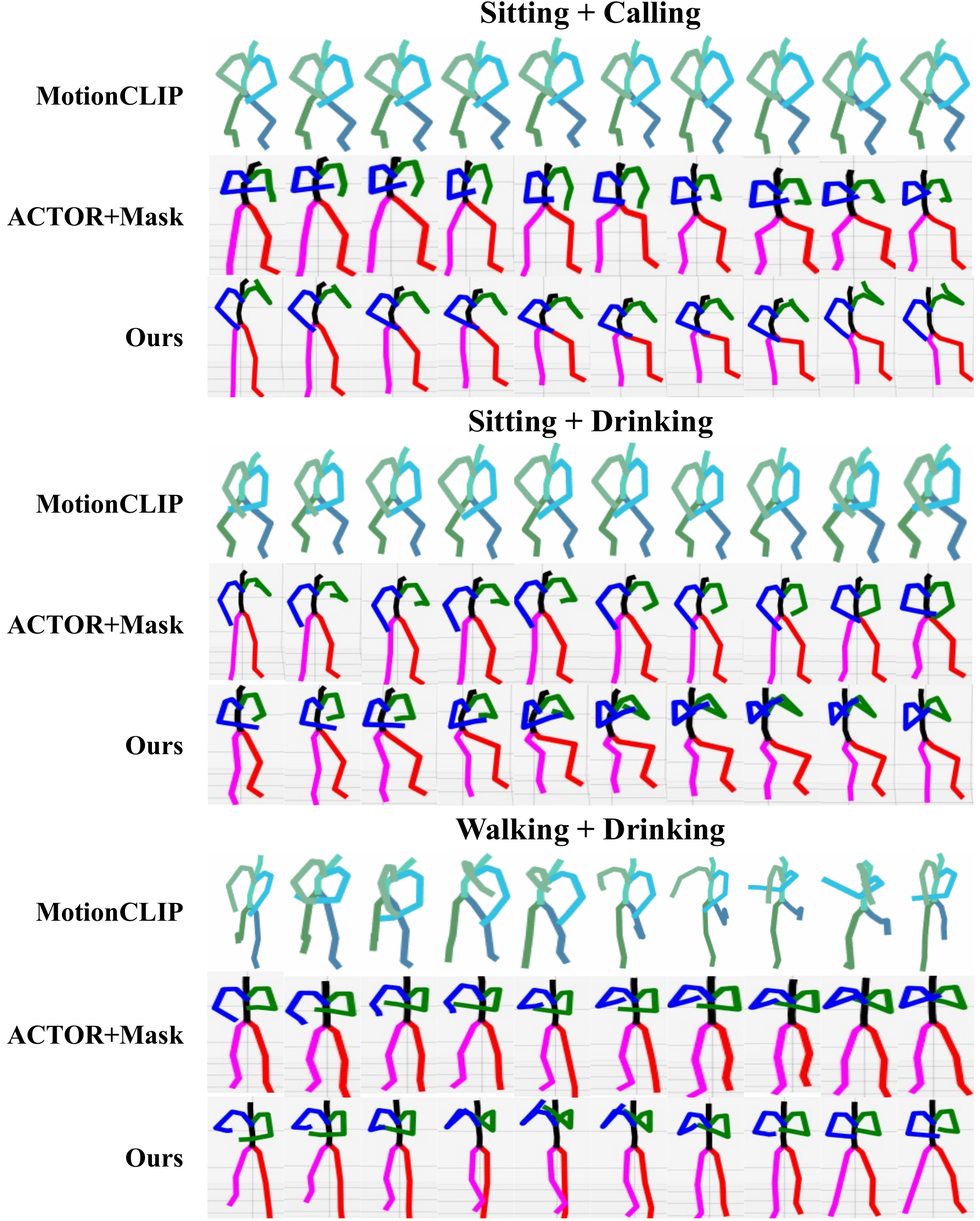}}
    \caption{ The qualitative comparisons  between our methods and other baseline methods. 
  }
    \label{fig:compare}
\end{figure}

Figure~\ref{fig:compare} compares the generations among the Pre-trained ACTOR~\cite{actor} + Mask, MotionCLIP~\cite{motionclip}, and ours. We notice that MotionCLIP does not generate actions that are easily distinguishable or adequately representative of the intended behaviors. For instance, the actions corresponding to "sitting + calling" and "sitting + drinking" are quite similar with limited variations, making it difficult to identify the distinct characteristics of calling or drinking in the animations.
On the other hand, ACTOR+Mask manages to generate reasonable composite actions, but these actions lack fluidity and some characteristic elements are not clearly defined. Taking "walking + drinking" as an example, the animation does not exhibit a clear walking motion. 
By contrast, our proposed method demonstrates an improved capacity to generate more discernible and smooth composite actions, highlighting the key features of each individual action. This comparison underscores the superiority of our approach in generating coherent and representative composite actions. Please note that we use a white background specifically for visualizations involving MotionCLIP to indicate that it was not trained on our dataset. However, it's important to emphasize that the choice of the background color is purely cosmetic and does not have any bearing on the quantitative performance of the model.

\begin{figure}[t]
    \centering
\centerline{\includegraphics[width=\linewidth]{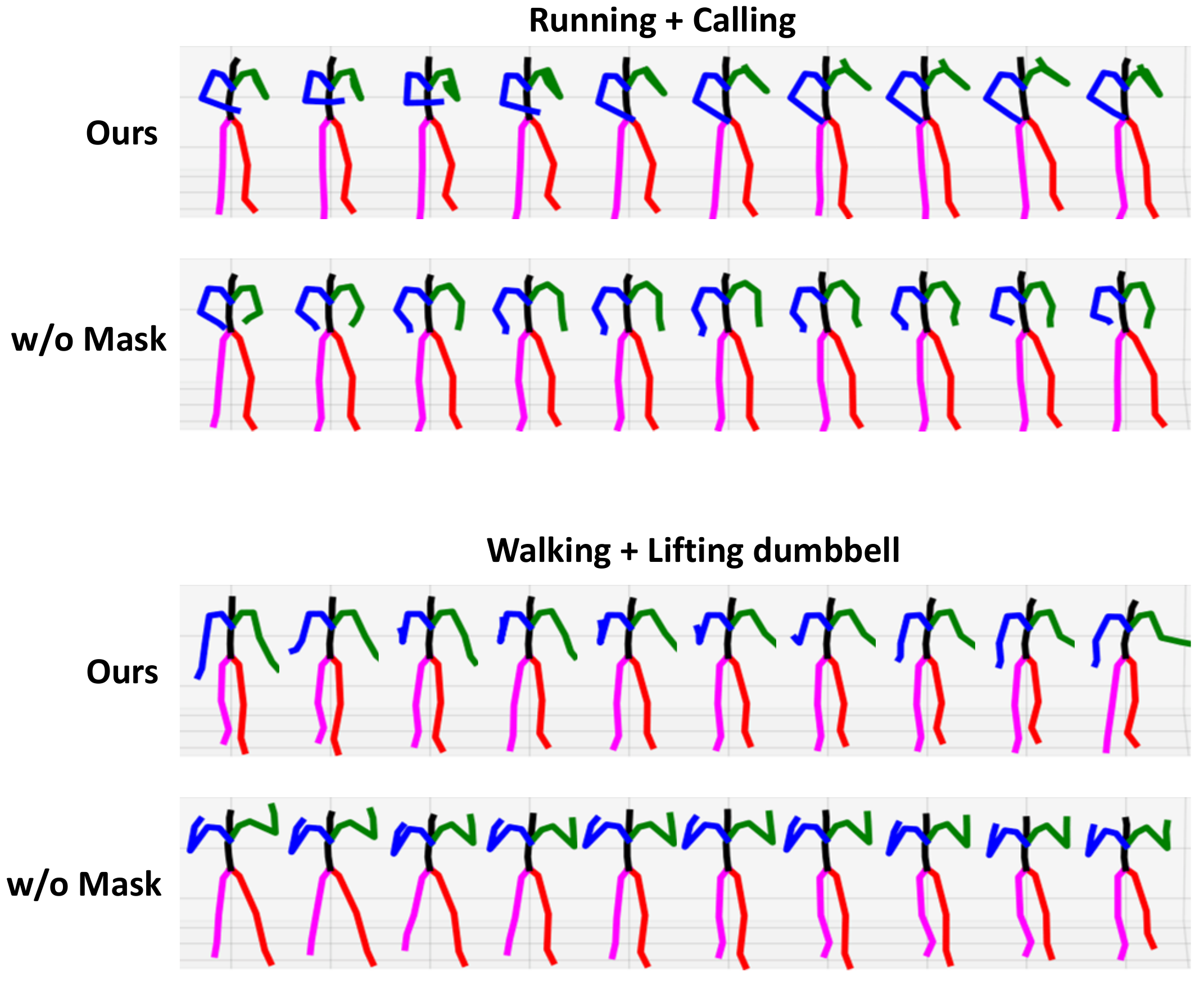}}
    \caption{ The qualitative comparisons between our methods and the one without motion energy for building attention.
  }
    \label{fig:compare_energy}
\end{figure}

In Section A.3.4 and Table A.1, we conducted a quantitative analysis of the influence of motion energy. In this addendum, we supplement this analysis with visual comparisons displayed in Figure~\ref{fig:compare_energy}. 
A cursory visual examination reveals that models without the application of motion energy may fail to capture the crucial characteristics of the sub-actions. For instance, in the "run + call" action, the model neglects the characteristic motion of making a phone call when motion energy is not used to emphasize the significant parts of the sub-action. This further demonstrates the critical role that motion energy plays in accurately capturing and preserving the distinctive characteristics of each sub-action in the final composed motion.

\end{document}